\documentclass{article}

    \PassOptionsToPackage{numbers, compress}{natbib}

\usepackage[preprint]{neurips_2026}

\usepackage{amsmath}
\usepackage{graphicx}
\usepackage[hidelinks,colorlinks=false]{hyperref}  
\usepackage[utf8]{inputenc} 
\usepackage[T1]{fontenc}    
\usepackage{hyperref}       
\usepackage{url}            
\usepackage{booktabs}       
\usepackage{amsfonts}       
\usepackage{nicefrac}       
\usepackage{microtype}      
\usepackage{xcolor}         
\usepackage{graphicx}
\usepackage{subfig}
\usepackage{multirow} 
\usepackage{enumitem}
\usepackage{wrapfig}
\usepackage{amsmath}
\usepackage{pifont}
\usepackage{algorithm}
\usepackage{algpseudocode}
\usepackage[normalem]{ulem}
\title{PARD-2: Target-Aligned Parallel Draft Model for Dual-Mode Speculative Decoding}

\author{
Zihao~An$^{1}$\thanks{Equal contribution.}\quad
Taichi~Liu$^{1,2}$\footnotemark[1]\quad
Ziqiong~Liu$^{1}$\quad
Dong~Li$^{1}$\quad
Ruofeng Liu$^{3}$ \quad
Emad~Barsoum$^{1}$
\\
${}^{1}$Advanced Micro Devices, Inc. \quad 
${}^{2}$Rutgers University \quad
${}^{3}$Michigan State University
\\
\{Zihao.An, Taichi.Liu, Ziqiong.Liu, d.li, Emad.Barsoum\}@amd.com, liuruofe@msu.edu
}

\begin{document}

\maketitle

\begin{abstract}
Speculative decoding accelerates Large Language Models (LLMs) inference by using a lightweight draft model to propose candidate tokens that are verified in parallel by the target model. However, existing draft model training objectives are not directly aligned with the inference-time goal of maximizing consecutive token acceptance. To address this issue, we reformulate the draft model optimization objective, shifting the focus from token prediction accuracy to the overall acceptance length.  In this paper, we build upon PARD to propose PARD-2, a dual-mode speculative decoding framework with Confidence-Adaptive Token (CAT) optimization. This approach adaptively reweights each token to better align with the verification process. Notably, PARD-2 enables a single draft model to support both target-dependent and target-independent modes. Experiments across diverse models and tasks demonstrate that PARD-2 achieves up to 6.94$\times$ lossless acceleration, surpassing EAGLE-3 by 1.9$\times$ and PARD by 1.3$\times$ on Llama3.1-8B. Our code is available at \url{https://github.com/AMD-AGI/PARD}.
\end{abstract}

\vspace{-2ex}
\section{Introduction}

As Large Language Models (LLMs) continue to advance, their strong performance has been accompanied by a rapid increase in model scale. While this scaling law has led to remarkable capability gains, it also makes auto-regressive decoding increasingly expensive at inference time.

Speculative Decoding (SD)~\citep{leviathan2023fast} has recently emerged as an effective approach to reducing LLMs inference latency. 
SD uses a lightweight draft model to propose multiple candidate tokens, which are then verified in parallel by the target model.
A promising line of work trains lightweight auto-regressive drafters conditioned on target-model features, including methods such as Medusa~\citep{cai2024medusa}, Hydra~\citep{ankner2024hydra}, and EAGLE-3~\citep{li2025eagle}, which achieve strong performance. However, sequential drafting still requires multiple forward passes, resulting in non-negligible latency~\citep{an2025pard, xiao2024parallelspec}. To further accelerate drafting, recent works explore parallel drafting to further reduce drafting latency: ParallelSpec~\citep{xiao2024parallelspec} trains a parallel drafter to generate multiple tokens in a single forward pass, PARD~\citep{an2025pard} adapts small auto-regressive models for parallel masked-token prediction, while DFlash~\citep{chen2026dflash} employs a small block diffusion model to generate draft tokens in parallel.

\begin{figure}[ht]
\centering
   \setlength{\abovecaptionskip}{2pt}
   \setlength{\belowcaptionskip}{2pt}
      \subfloat[Llama3.1-8B]{ \includegraphics[width=0.48\columnwidth]{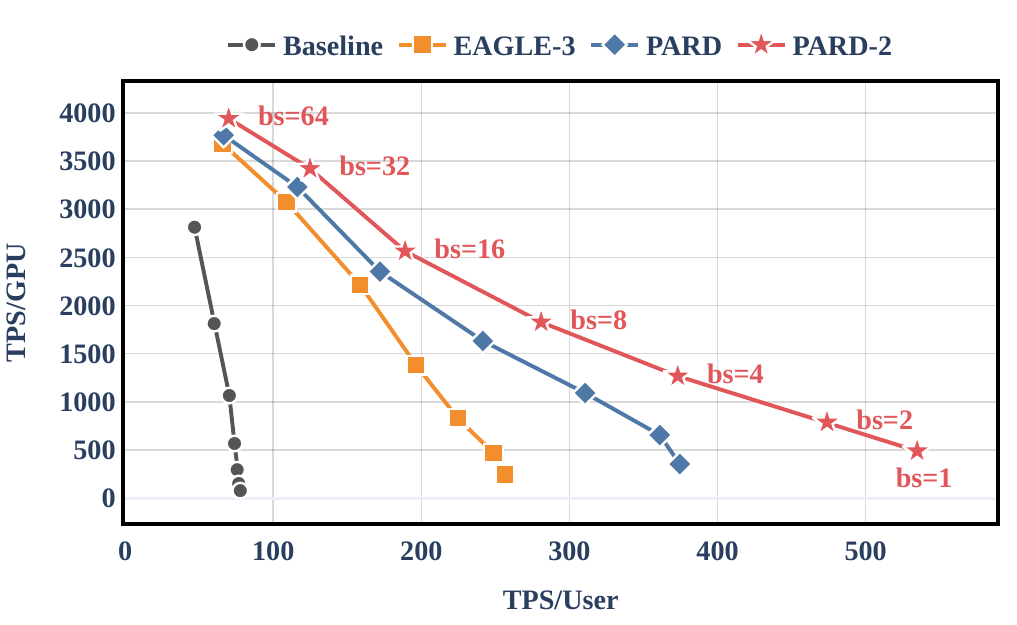}
      \label{fig:frame1}}
      \hfill
      \subfloat[Qwen3-8B]{ \includegraphics[width=0.48\columnwidth]{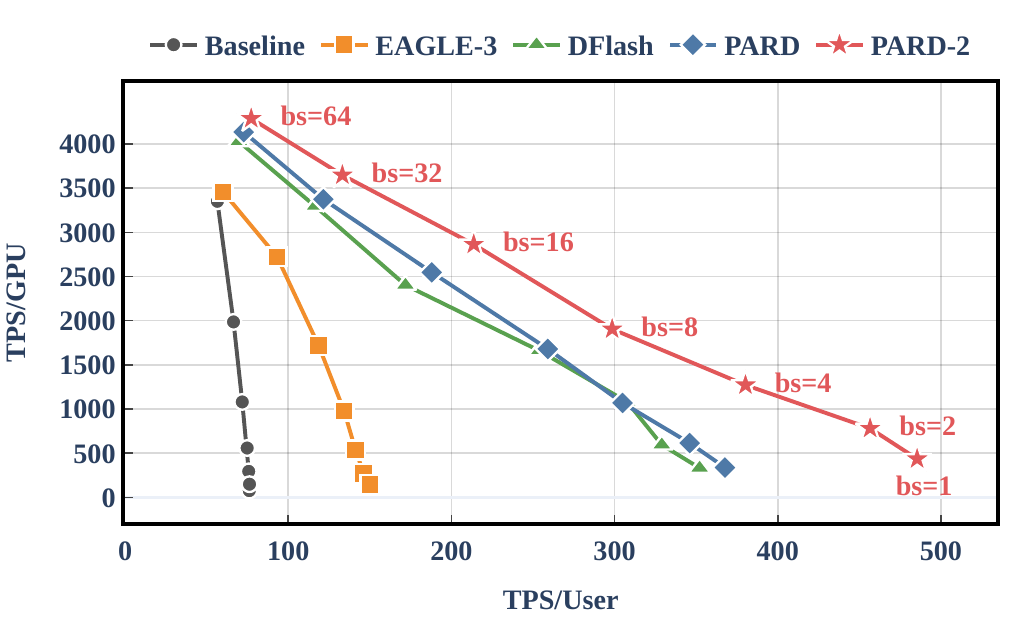} \label{fig:frame3}} 
      
\caption{\textbf{Throughput and Latency Trade-offs on vLLM.}  PARD-2 consistently achieves a superior Pareto frontier across various batch sizes (1 to 64) on both (a) Llama-3.1-8B and (b) Qwen-3-8B.}
\label{fig:bs}
\end{figure}

However, a common assumption underlying speculative decoding is that all draft positions should be learned equally in training time, which is suboptimal for training convergence and acceptance length~\citep{liu2026dart, chen2026dflash}. Unlike standard language modeling, where the objective is to improve token prediction accuracy uniformly, speculative decoding is ultimately concerned with how many drafted tokens can be accepted by the target model.
Our experiments reveal a positional bias in parallel speculative decoding: as illustrated in Figure~\ref{fig:observation}(a), tokens at subsequent draft positions exhibit consistently lower acceptance rates. As the draft length increases, the acceptance rate often struggles to persist, limiting the practical speedup that parallel drafting can provide. This observation suggests an \emph{inherent limitation of uniformly optimizing all positions}. 
While recent approaches such as DFlash~\citep{chen2026dflash} and DART~\citep{liu2026dart} mitigate this issue with position-aware decaying weights, their weights are fixed and primarily position-dependent. We observe that a token's acceptance is determined not solely by the accuracy of the current token, but is heavily bottlenecked by the quality of the entire prefix. This indicates that acceptance is jointly determined by the current token and its prefix context. Therefore, an approach that jointly considers both of these two factors provides a more effective way to improve acceptance length and decoding efficiency.

In this paper, we introduce PARD-2, a dual-mode speculative decoding framework to mitigate the degradation in acceptance rate. We propose Confidence-Adaptive Token (CAT) optimization, which assigns token-level, context-dependent confidence scores to better align the training objective with the inference-time goal of maximizing consecutive token acceptance in speculative decoding. Specifically, CAT dynamically reweights token-level objectives based on a context-dependent confidence score, which is computed as the cumulative product of the target model's confidence across all preceding tokens in the prefix. This design encourages the drafter to maximize the expected acceptance length.

In addition to optimizing acceptance length, PARD-2 further addresses the target dependency of existing speculative decoding methods. Most speculative decoding methods are target-dependent~\citep{li2025eagle, chen2026dflash}, requiring training a new draft model from scratch for each target model. Building upon PARD, PARD-2 is the first to enable a single draft model to dynamically switch between target-dependent and target-independent modes during inference. Unlike EAGLE-3 and DFlash, which require grafted layers, PARD-2 maintains a standalone architecture, achieving this flexibility without structural overhead. It applies stochastic gating to control the injection of target hidden states during training. As a result, the same draft model can operate in a target-dependent mode for maximum acceleration, while also supporting a target-independent mode that generalizes across a family of target models. 

To summarize, our key contributions include:
\begin{itemize}
    \item We propose PARD-2, a dual-mode speculative decoding framework that supports both target-dependent and target-independent settings. To the best of our knowledge, this is the first work to unify these paradigms within a single draft model. Stochastic gating injects target hidden states during training, enabling peak acceleration via target-dependent optimization while maintaining universal compatibility with an entire model family.     
   
    \item We revisit the fundamental objective of speculative decoding and demonstrate that its primary challenge is maximizing the acceptance of consecutive token spans. To this end, we propose a novel optimization strategy CAT. Conditioned on the preceding prefix, CAT adaptively reweights its focus on individual tokens guided by the target model’s context-dependent confidence scores, thereby significantly improving both prediction and distillation efficiency.
    \item We conduct extensive experiments across diverse models and benchmarks, including a practical validation of PARD-2 within the vLLM framework. Our results show that PARD-2 achieves an average speedup of 1.3× over PARD and up to 6.94× acceleration over the autoregressive baseline. Furthermore, it delivers the highest throughput under high-concurrency settings, demonstrating exceptional practical value for real-world deployment.

\end{itemize}

\section{Preliminaries}

\subsection{Speculative Decoding}
Speculative decoding is a lossless decoding strategy for accelerating LLM inference. Instead of generating each token solely with the target model $\boldsymbol{\theta}_{\mathrm{target}}$, it introduces a smaller and faster draft model $\boldsymbol{\theta}_{\mathrm{draft}}$ to propose multiple candidate tokens in advance, which are then verified by the target model in parallel. This design reduces the number of expensive target model decoding steps while preserving the exact output distribution of the target model.

\begin{figure}[!t]
\centering
   \setlength{\abovecaptionskip}{2pt}
   \setlength{\belowcaptionskip}{2pt}
      \subfloat[Position-wise acceptance ratio and acceptance length]{ \includegraphics[width=0.485\columnwidth]{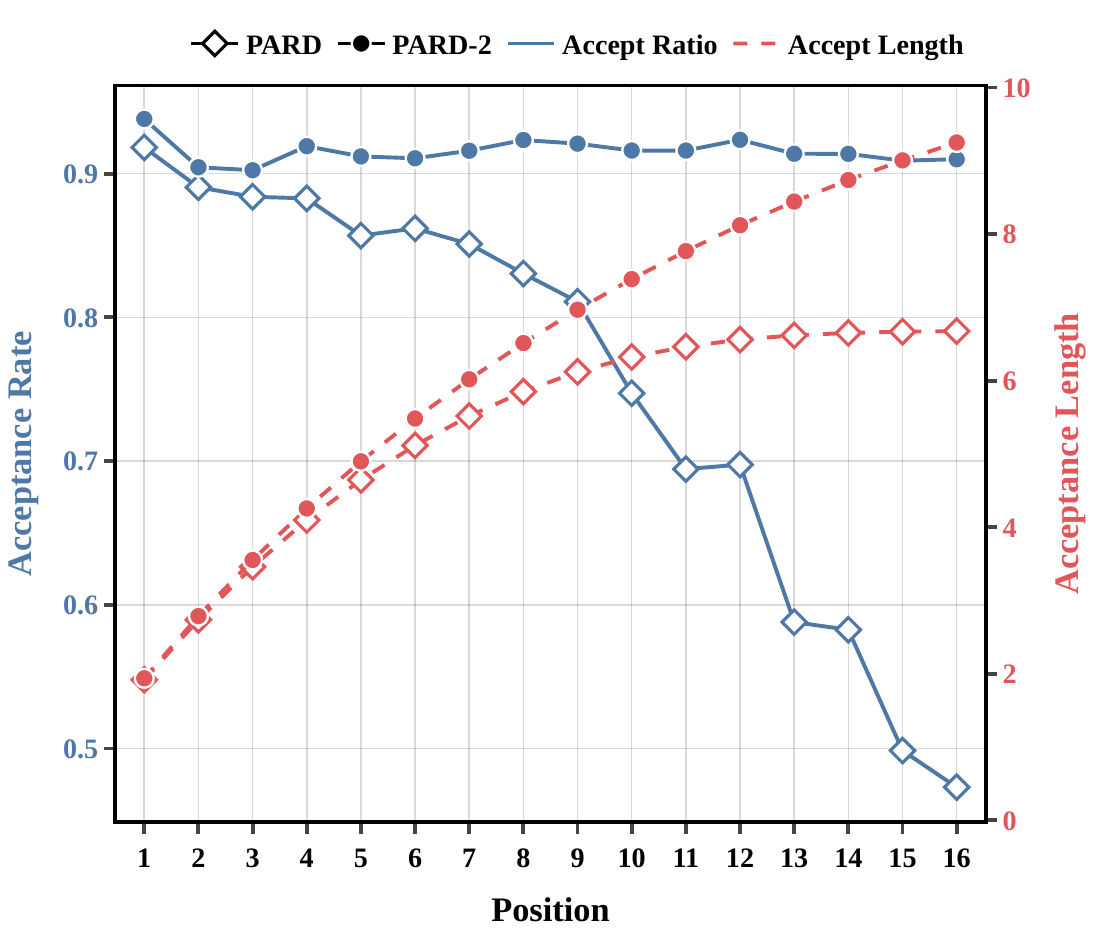}
      \label{fig:infer design}}
      \hfill
      \subfloat[Target Model's Confidence vs Accpetance Rate]{ \includegraphics[width=0.48
      \columnwidth]{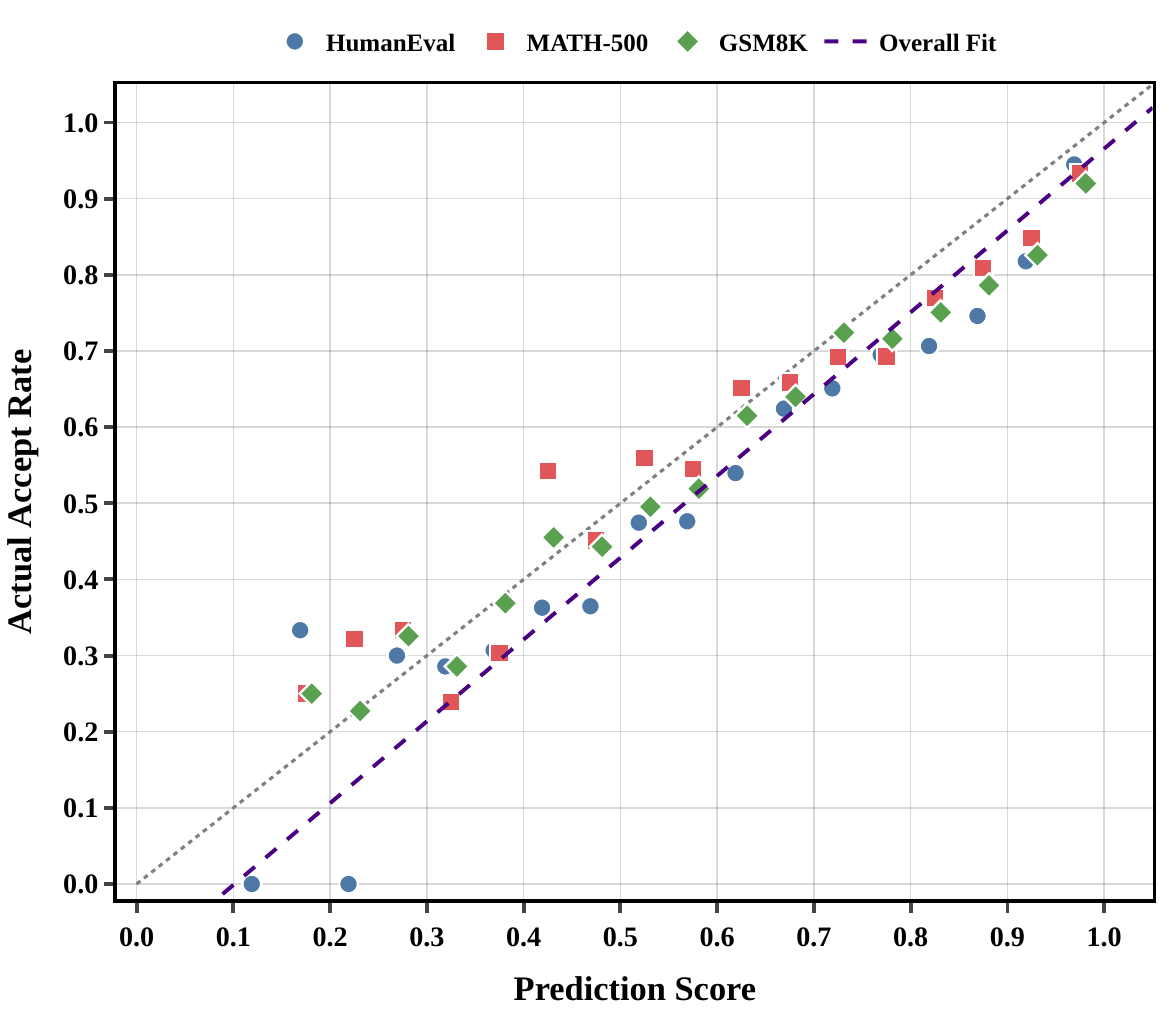} \label{fig:position-aware}} 
      
\caption{
\textbf{Acceptance behavior of Llama3.1-8B.}
(a) On the HumanEval benchmark, PARD-2 achieves higher acceptance rates and longer acceptance length than PARD across token positions, mitigating distant-position degradation.
(b) Target-model confidence scores strongly correlate with actual acceptance rates, supporting their use as a proxy for token-level acceptance.
}
\label{fig:observation}
\end{figure}

Formally, given a prefix $X=(x_0,\ldots,x_{n-1})$, speculative sampling uses a lightweight auto-regressive draft model $\boldsymbol{\theta}_{\mathrm{draft}}$ to propose a length of $K$ tokens, denoted by $\tilde{Y}=(\tilde{y}_n,\ldots,\tilde{y}_{n+K-1})$. The proposal probability distribution factorizes as
\begin{equation}
P(\tilde{Y}\mid X;\boldsymbol{\theta}_{\mathrm{draft}})
=
\prod_{k=0}^{K-1}
P\!\left(
\tilde{y}_{n+k}\mid
x_0,\ldots,x_{n-1},
\tilde{y}_n,\ldots,\tilde{y}_{n+k-1};
\boldsymbol{\theta}_{\mathrm{draft}}
\right).
\label{eq:ar_draft}
\end{equation}
For position $n+k$, let
$
p_k(y)=P(y\mid x_0,\ldots,x_{n-1},\tilde{y}_n,\ldots,\tilde{y}_{n+k-1};\boldsymbol{\theta}_{\mathrm{target}})
$
and
$
q_k(y)=P(y\mid x_0,\ldots,x_{n-1},\tilde{y}_n,\ldots,\tilde{y}_{n+k-1};\boldsymbol{\theta}_{\mathrm{draft}})
$
denote the target and draft conditional probabilities, respectively. Under speculative sampling, the draft token $\tilde{y}_{n+k}$ is accepted with probability
\begin{equation}
a_k
=
\min\!\left(
1,\,
\frac{p_k(\tilde{y}_{n+k})}{q_k(\tilde{y}_{n+k})}
\right),
\qquad k=0,\ldots,K-1.
\label{eq:accept_prob}
\end{equation}
Ignoring the bonus token, the probability that the first $k+1$ draft tokens are all accepted is $\prod_{j=0}^{k} a_j$. Hence, the expected acceptance length $L$ is
\begin{equation}
\mathbb{E}[L\mid X,\tilde{Y}]
=
\sum_{k=0}^{K-1}\prod_{j=0}^{k} a_j.
\label{eq:expected_len}
\end{equation}
The target model accepts the longest valid prefix and, upon the first rejection, samples a correction token from the residual distribution, preserving exact equivalence to sampling from the target model.

\subsection{Parallel Draft Models}

Although speculative decoding significantly accelerates LLM inference, its drafting stage remains sequential, requiring $K$ sequentially dependent predictions to generate $K$ draft tokens. This sequential latency can still limit the end-to-end speedup. To address this issue, recent work has explored parallel draft models that predict multiple tokens simultaneously. DiffuSpec~\citep{li2025diffuspec} and DFlash~\citep{chen2026dflash} adopt diffusion-based drafters that generate tokens through iterative denoising. To better match the auto-regressive architecture of the target model, PARD~\citep{an2025pard} retains an auto-regressive backbone and introduces masked placeholders, enabling parallel masked-token prediction in a single forward pass.

In particular, PARD introduces a special mask token $m$ and predicts each future token conditioned only on the prefix and preceding mask placeholders. Its draft probability distribution is
\begin{equation}
P(\tilde{Y}\mid X;\boldsymbol{\theta}_{\mathrm{PARD}})
=
\prod_{k=0}^{K-1}
P\!\left(
\tilde{y}_{n+k}\mid
x_0,\ldots,x_{n-1},
m_n,\ldots,m_{n+k-1};
\boldsymbol{\theta}_{\mathrm{PARD}}
\right).
\label{eq:pard_draft}
\end{equation}
Because each position depends only on the prefix and mask tokens, all $K$ predictions can be computed in a single forward pass. 
This approach not only substantially reduces drafting latency but also ensures target independence, enabling the drafter to be reusable across a family of target models.

Given the ground-truth $Y=(y_n,\ldots,y_{n+K-1})$, PARD is trained with the cross-entropy loss
\begin{equation}
\mathcal{L}_{\mathrm{PARD}}
=
-\frac{1}{K}
\sum_{k=0}^{K-1}
\log
P\!\left(
y_{n+k}\mid
x_0,\ldots,x_{n-1},
m_n,\ldots,m_{n+k-1};
\boldsymbol{\theta}_{\mathrm{PARD}}
\right).
\label{eq:pard_ce}
\end{equation}

\begin{figure*}[!t]
\centering
\setlength{\abovecaptionskip}{2pt}
\setlength{\belowcaptionskip}{2pt}
\includegraphics[width=\textwidth]{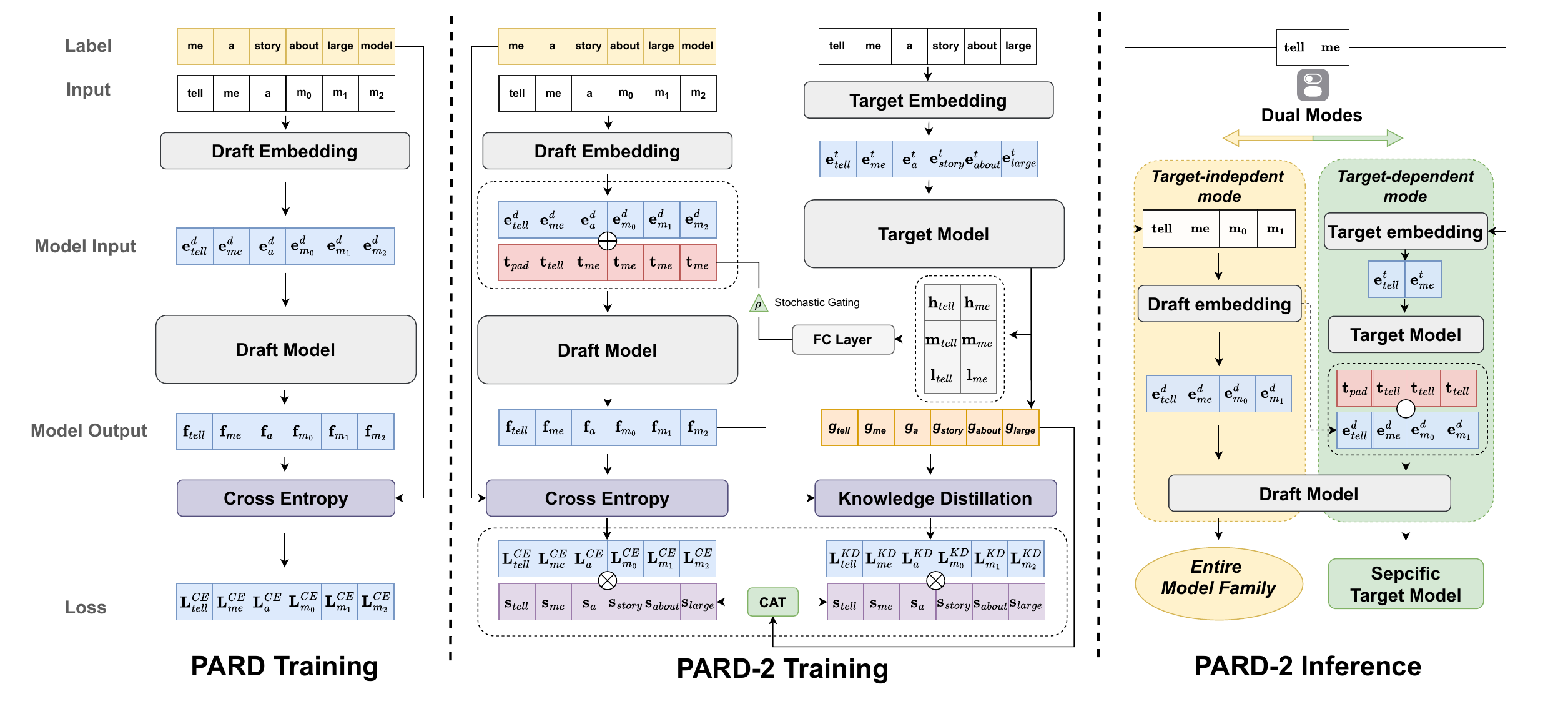}
\caption{\textbf{Overview of PARD-2.} The training (mid) and inference (right) designs of PARD-2. Compared to PARD (left), PARD-2 integrates CAT optimization, target hidden features, and knowledge distillation. PARD-2 supports flexible switching between target dependent and independent modes.}
\label{fig:train_and_infer}
\end{figure*}

\section{Method}


\subsection{Observation}
The draft length $K$ is a key design choice for parallel draft models. To study its effect, we train PARD with two draft lengths, $K=8$ and $K=16$. As shown in Table~\ref{table:ab_main}, increasing $K$ yields little improvement and can even degrade performance across several benchmarks. This observation contradicts the common intuition that a longer draft length should naturally translate to a greater acceptance length and enhanced decoding efficiency. To understand this phenomenon, we analyze the verification mechanism of speculative decoding. Because the target model evaluates candidate tokens strictly in order, the acceptance of any subsequent token heavily relies on the successful verification of all its predecessors. Let $a_j$ denote the marginal probability that the target model accepts the $j$-th draft token. Eq.~\eqref{eq:expected_len} can be decomposed by position as
\begin{equation}
\mathbb{E}[L\mid X,\tilde{Y}]
=
\sum_{k=0}^{K-1}\prod_{j=0}^{k} a_j
=
\sum_{k=0}^{K-1}
\left(\prod_{j=0}^{k-1} a_j\right)a_k.
\label{eq:expected_len_decompose}
\end{equation}
This decomposition reveals two key factors that govern whether the token at position $k$ is accepted. The first factor, $\prod_{j=0}^{k-1} a_j$, is the probability that all previous draft tokens are accepted.
The second factor $a_k$ is the probability that the current token is accepted once position $k$ is reached.

We then define the first factor as $s_k$:
\begin{equation}
s_k
:=
\prod_{j=0}^{k-1} a_j,
\qquad s_0:=1.
\label{eq:pis}
\end{equation}
The term $s_k$ is a prerequisite for the $k$-th token to contribute to the acceptance length and can therefore be interpreted as the importance of that token with respect to acceleration. With this notation,
\begin{equation}
\mathbb{E}[L\mid X,\tilde{Y}]
=
\sum_{k=0}^{K-1} s_k a_k.
\label{eq:expected_len_pis}
\end{equation}

The second factor, $a_k$, reflects the local quality of the draft prediction at position $k$. Since the token-level training objective of the draft model aims to improve prediction quality at each position, it is naturally related to increasing $a_k$. This insight motivates us to reweight the per-token training objectives by $s_k$, assigning higher importance to tokens situated on highly probable accepted prefixes.

\subsection{Confidence-Adaptive Token Optimization Strategy}
Motivated by the above observations, we assign adaptive weights to individual tokens during training, thereby better aligning the optimization objective with the speculative decoding goal of maximizing acceptance length. However, the true acceptance rate $a_k$ during training is intractable, as it inherently depends on the dynamic interaction between the draft and target models during the verification phase. 

Inspired by EAGLE-2~\citep{li2024eagle2fasterinferencelanguage} and CAPE~\citep{du2024glide}, we investigate the relationship between the target model's confidence and the token acceptance rate across multiple benchmarks, including HumanEval, GSM8K, and Math-500. As illustrated in Figure~\ref{fig:observation}(b), the target model's confidence exhibits a strong positive correlation with the empirical acceptance rate. Consequently, we can leverage the target model's confidence scores as a reliable proxy for the expected token acceptance rate, enabling adaptive token-level weighting.

Building upon this empirical finding, we approximate the actual acceptance probability using the target model's confidence on the corresponding ground-truth token $y_{n+k}$ conditioned on its prefix:
\begin{equation}
\hat{c}_{k}
:=
P\!\left(
y_{n+k}\mid
x_0,\ldots,x_{n-1},
y_n,\ldots,y_{n+k-1};
\boldsymbol{\theta}_{\mathrm{target}}
\right).
\label{eq:confidence_proxy}
\end{equation}
We then estimate the importance of each token by computing the cumulative product of the target confidences along the prefix:
\begin{equation}
\hat{s}_k
:=
\prod_{j=0}^{k-1}\hat{c}_j,
\qquad \hat{s}_0:=1.
\label{eq:pis_estimate}
\end{equation}
Here, $\hat{s}_k$ approximates the probability that position $k$ is reached during verification.
Using $\hat{s}_k$ as a stop-gradient weight, we obtain the final PARD-2 training objective:
\begin{equation}
\mathcal{L}_{\mathrm{PARD\text{-}2}}
=
-\frac{1}{K}
\sum_{k=0}^{K-1}
\hat{s}_k
\log
P\!\left(
y_{n+k}\mid
x_0,\ldots,x_{n-1},
m_n,\ldots,m_{n+k-1};
\boldsymbol{\theta}_{\mathrm{PARD\text{-}2}}
\right).
\label{eq:pard2_loss}
\end{equation}
Unlike the uniform loss in Eq.~\eqref{eq:pard_ce}, Eq.~\eqref{eq:pard2_loss} adaptively prioritizes tokens that are likely to contribute to the final accepted prefix and reduces the influence of distant positions that are rarely reached during speculative verification. As a result, the resulting objective better matches the inference-time acceleration goal of speculative decoding.
\subsection{PARD-2 Training}

During training, in addition to assigning different importance weights to each token-level loss to better optimize the speculative decoding objective, we further adapt the draft model to align with the target model, enabling it to generate highly compatible proposals.

\textbf{Stochastic Gating for Target Features.} As illustrated in Figure~\ref{fig:train_and_infer}, given an input prompt $X$, we extract hidden representations from multiple layers of the target model, denoted by $l$, $m$, and $h$, corresponding to low-, middle-, and high-level features, respectively. These hidden states are fused into a compact target-context feature $t = \mathrm{Proj}([l; m; h])$, where $[\cdot ; \cdot]$ denotes concatenation and $\mathrm{Proj}(\cdot)$ is a lightweight projection module. To improve training efficiency, we further process the draft-model input with Conditional Drop Token (COD)~\citep{an2025pard}, which selectively drops conditional tokens during training. The fused target hidden feature $t$ is then injected into the draft model by adding it to the draft-model input embeddings $e^d$. In this way, the draft model can leverage target-side context during drafting, leading to better alignment with the target-model distribution.

Moreover, to achieve target independence, we do not inject target-context features for every training instance. Instead, we stochastically inject them during training: \(e^{d'} = e^d + \xi \cdot t\), \(\xi \sim \mathrm{Bernoulli}(1-\rho)\). Thus, \(e^{d'}=e^d\) with probability \(\rho\), and \(e^{d'}=e^d+t\) otherwise. This design reduces target-side dependence, enabling a single drafter to serve the entire model family.

\textbf{Training Loss Function.}
To further improve draft-target alignment, we augment the supervised training objective with knowledge distillation from the target model. 
Without loss of generality, let $x_{\le n}$ denote the current prefix and $x_{n+k}$ denote the $k$-th future token to be predicted. 
Unlike a position-only weight $\hat{s}_{k}$, the acceptance probability of a future token depends not only on its relative position $k$, but also on the current prefix $x_{\le n}$. 
Therefore, we denote the estimated token-level acceptance weight as $\hat{s}_{n,k}$, which captures the expected acceptance likelihood of token $x_{n+k}$ conditioned on prefix $x_{\le n}$. Specifically, for the $k$-th future position, our objective is formulated as a weighted sum of the standard cross-entropy loss $\mathcal{L}^{\mathrm{CE}}$ and the distillation loss $\mathcal{L}^{\mathrm{KD}}$:
\begin{equation}
\mathcal{L}_{k}
=
\sum_{n=1}^{N-k}
\hat{s}_{n,k}
\left(
\beta \mathcal{L}^{\mathrm{CE}}_{n,k}
+
\mathcal{L}^{\mathrm{KD}}_{n,k}
\right),
\label{eq:loss_k}
\end{equation}
where $\beta$ balances the supervised and distillation terms. 
Here, $\mathcal{L}^{\mathrm{CE}}_{n,k}$ and $\mathcal{L}^{\mathrm{KD}}_{n,k}$ are computed for token $x_{n+k}$ given prefix $x_{\le n}$, and $\hat{s}_{n,k}$ is the confidence-adaptive weight assigned to this token. 

\subsection{Dual-mode inference}
During inference, our framework supports two modes: target-dependent and target-independent. As illustrated in Figure~\ref{fig:train_and_infer}, during the standard prefilling phase, we extract hidden representations across multiple layers of the target model. To maintain consistency between training and inference when utilizing the target model's features, we specifically extract the hidden state corresponding to the last token position of the prompt to fuse with the mask token. This design ensures that the draft model's conditioning signal remains aligned with the sequential dependency observed during the training phase. In target-dependent mode, PARD-2 maximizes alignment by exploiting target hidden features to achieve peak acceleration. In target-independent mode, it maintains broad compatibility across the entire target model family without retraining. 
Notably, both modes are supported by the same single draft model during inference, requiring no architectural changes or additional parameter fine-tuning.

\begin{table*}[ht]
  \centering
  \scriptsize
  \renewcommand{\arraystretch}{0.95}
  \setlength{\tabcolsep}{4pt}
  \caption{Target-dependent comparison of speedup ratios and average acceptance lengths $\tau$ across different methods. Q3 represents Qwen3 model family and L3 represents Llama3 model family. Values in parentheses denote the inference draft length.}
  \resizebox{\linewidth}{!}{
  \begin{tabular}{@{}l l@{\hspace{6pt}}r *{10}{c}@{}}
    \toprule
    \textbf{Target} & \multicolumn{2}{l}{\textbf{Method}} &
    \multicolumn{2}{c}{\textbf{HumanEval}} &
    \multicolumn{2}{c}{\textbf{GSM8K}} &
    \multicolumn{2}{c}{\textbf{Math-500}} &
    \multicolumn{2}{c}{\textbf{MT-Bench}} &
    \multicolumn{2}{c}{\textbf{Average}} \\
    \cmidrule(lr){4-5} \cmidrule(lr){6-7} \cmidrule(lr){8-9} \cmidrule(lr){10-11} \cmidrule(lr){12-13}
    \multicolumn{3}{c}{\textit{Temperature = 0}} &
    SpeedUp & $\tau$ & SpeedUp & $\tau$ & SpeedUp & $\tau$ & SpeedUp & $\tau$ & SpeedUp & $\tau$ \\
    \midrule
    
    Q3 8B
      & AR      &                    & 1.00 & 1.00 & 1.00 & 1.00 & 1.00 & 1.00 & 1.00 & 1.00 & 1.00 & 1.00 \\
      & EAGLE-3 & {\color{gray}(8)}  & 2.11 & 2.68 & 1.98 & 2.44 & 1.90 & 2.32 & 1.71 & 2.05 & 1.93 & 2.37 \\
      & DFlash  & {\color{gray}(16)} & 5.66 & 6.94 & 4.64 & 5.39 & 5.59 & 6.88 & 2.57 & 3.10 & 4.61 & 5.58 \\
      & PARD    & {\color{gray}(16)} & 5.02 & 6.33 & 4.91 & 6.01 & 4.96 & 6.12 & 2.68 & 3.79 & 4.39 & 5.56 \\
      & PARD-2  & {\color{gray}(16)} & \textbf{6.75} & \textbf{8.09} & \textbf{6.44} & \textbf{7.68} & \textbf{6.87} & \textbf{8.27} & \textbf{3.16} & \textbf{3.89} & \textbf{5.81} & \textbf{6.98} \\
    \midrule
    Q3 14B
      & AR      &                    & 1.00 & 1.00 & 1.00 & 1.00 & 1.00 & 1.00 & 1.00 & 1.00 & 1.00 & 1.00 \\
      & EAGLE-3 & {\color{gray}(8)}  & 2.32 & 2.86 & 2.10 & 2.50 & 2.09 & 2.49 & 1.72 & 1.97 & 2.06 & 2.46 \\
      & PARD    & {\color{gray}(16)} & 5.16 & 5.25 & 5.10 & 6.09 & 4.98 & 5.95 & 2.78 & 3.26 & 4.51 & 5.14 \\
      & PARD-2 & {\color{gray}(16)} & \textbf{6.74} & \textbf{7.95} & \textbf{6.40} & \textbf{7.65} & \textbf{6.73} & \textbf{8.07} & \textbf{3.37} & \textbf{3.96} & \textbf{5.81} & \textbf{6.91} \\
    \midrule
    L3.1 8B
      & AR      &                    & 1.00 & 1.00 & 1.00 & 1.00 & 1.00 & 1.00 & 1.00 & 1.00 & 1.00 & 1.00 \\
      & EAGLE-3 & {\color{gray}(8)}  & 3.34 & 4.55 & 2.80 & 3.97 & 2.37 & 3.42 & 2.54 & 3.60 & 2.76 & 3.89 \\
      & PARD    & {\color{gray}(16)} & 5.09 & 6.71 & 4.20 & 5.43 & 3.84 & 5.08 & 2.85 & 3.57 & 4.00 & 5.20 \\
      & PARD-2  & {\color{gray}(16)} & \textbf{6.94} & \textbf{9.19} & \textbf{5.21} & \textbf{6.98} & \textbf{5.36} & \textbf{7.40} & \textbf{3.25} & \textbf{4.22} & \textbf{5.19} & \textbf{6.95} \\
    \bottomrule
  \end{tabular}
  }
  \label{tab:Target-Dependent Mode}
\end{table*}

\begin{table*}[ht]
  \centering
  \scriptsize
  \renewcommand{\arraystretch}{0.95}
  \setlength{\tabcolsep}{4pt}
  \caption{Target-independent Performance Comparison. Values in parentheses denote the inference draft length. All experiments are evaluated on the same draft model.}
  \resizebox{\linewidth}{!}{
  \begin{tabular}{@{}l l@{\hspace{6pt}}r *{10}{c}@{}}
    \toprule
    \textbf{Target} & \multicolumn{2}{l}{\textbf{Method}} &
    \multicolumn{2}{c}{\textbf{HumanEval}} &
    \multicolumn{2}{c}{\textbf{GSM8K}} &
    \multicolumn{2}{c}{\textbf{Math-500}} &
    \multicolumn{2}{c}{\textbf{MT-Bench}} &
    \multicolumn{2}{c}{\textbf{Average}} \\
    \cmidrule(lr){4-5} \cmidrule(lr){6-7} \cmidrule(lr){8-9} \cmidrule(lr){10-11} \cmidrule(lr){12-13}
    \multicolumn{3}{c}{\textit{Temperature = 0}} &
    SpeedUp & $\tau$ & SpeedUp & $\tau$ & SpeedUp & $\tau$ & SpeedUp & $\tau$ & SpeedUp & $\tau$ \\
    \midrule
    
    Q3 8B
      & AR            &                    & 1.00 & 1.00 & 1.00 & 1.00 & 1.00 & 1.00 & 1.00 & 1.00 & 1.00 & 1.00 \\
      & PARD    & {\color{gray}(16)} & 5.02 & 6.33 & 4.91 & 6.01 & 4.96 & 6.12 & 2.68 & 3.79 & 4.39 & 5.56 \\
      & PARD-2        & {\color{gray}(16)} & \textbf{5.71} & \textbf{7.38} & \textbf{5.20} & \textbf{6.56} & \textbf{5.63} & \textbf{7.23} & \textbf{2.74} & \textbf{3.39} & \textbf{4.82} & \textbf{6.14} \\
    \midrule
    Q3 14B
      & AR            &                    & 1.00 & 1.00 & 1.00 & 1.00 & 1.00 & 1.00 & 1.00 & 1.00 & 1.00 & 1.00 \\
      & PARD    & {\color{gray}(16)} & 5.16 & 5.25 & 5.10 & 6.09 & 4.98 & 5.95 & 2.78 & 3.26 & 4.51 & 5.14 \\
      & PARD-2        & {\color{gray}(16)} & \textbf{5.63} & \textbf{7.16} & \textbf{5.28} & \textbf{6.61} & \textbf{5.50} & \textbf{6.97} & \textbf{2.76} & \textbf{3.37} & \textbf{4.79} & \textbf{6.03} \\
    \midrule
    Q3 32B
      & AR            &                    & 1.00 & 1.00 & 1.00 & 1.00 & 1.00 & 1.00 & 1.00 & 1.00 & 1.00 & 1.00 \\
      & PARD          & {\color{gray}(16)} & 5.18 & 6.46 & 4.90 & 5.99 & 4.76 & 5.83 & 2.64 & 3.05 & 4.37 & 5.33 \\
      & PARD-2        & {\color{gray}(16)} & \textbf{5.71} & \textbf{7.12} & \textbf{5.10} & \textbf{6.24} & \textbf{5.25} & \textbf{6.47} & \textbf{2.67} & \textbf{3.15} & \textbf{4.68} & \textbf{5.75} \\
    \bottomrule
  \end{tabular}
  }
  \label{tab:Target-Independent Mode}
\end{table*}

\section{Experiments}
\subsection{EXPERIMENTAL SETUP}
\textbf{Models.} We evaluate PARD-2 primarily on the Llama3~\citep{llama3_2024} and Qwen3~\citep{yang2025qwen3} model families. To demonstrate performance via the target-dependent mode, we specifically train and evaluate PARD-2 on Llama-3.1-8B, Qwen3-8B, and Qwen3-14B. Furthermore, to highlight its zero-shot transferability, we conduct extensive target-independent experiments primarily on the Qwen3 family, demonstrating that a single drafter can seamlessly generalize to accelerate other target models within the same series.

\textbf{Datasets and Benchmarks.} PARD-2 is trained on a moderately expanded version of the dataset used in PARD. Specifically, we retain Magpie~\citep{xu2024magpie} and Evol-CodeAlpaca~\citep{luo2023wizardcoder}, and additionally include samples from Nemotron-v2~\citep{basant2025nvidia} and Nemotron-v3~\citep{blakeman2025nemotron}. We evaluate the generalizability of our approach across diverse benchmarks, including HumanEval~\citep{chen2021evaluating} for code generation, MATH-500~\citep{lightman2023lets} and GSM8K~\citep{cobbe2021training} for mathematical reasoning, and MT-Bench~\citep{zheng2023judging} for multi-turn dialogue.

\textbf{Metrics.} PARD-2 is a lossless acceleration method that preserves the original target model and exact acceptance rule. Therefore, we focus on acceleration performance and report the following metrics:
\begin{itemize} [topsep=0pt, parsep=0pt, itemsep=2pt, partopsep=0pt, leftmargin=10pt]
    \item \textbf{Speedup}: The acceleration ratio over vanilla auto-regressive decoding.
    \item \textbf{Acceptance Length $\tau$}: the average number of draft tokens accepted in each verification.
    \item \textbf{Tokens Per Second}: The number of tokens generated per second in real-world scenarios.
\end{itemize}

\textbf{Implementation Details.} For training, we extract target hidden features from 4 layers of the target model. The draft model is trained on AMD MI300X GPUs, utilizing a batch size of 64 and a draft length of $K=16$. We set $\rho=0.1$ and the loss weighting coefficient $\beta = 0.1$. For inference, all throughput evaluations are implemented based on the vLLM framework. To ensure a fair comparison, tree-based decoding is explicitly disabled across all methods. Unless otherwise specified, all evaluation experiments are conducted on NVIDIA A100-40GB GPUs. We employ a tensor parallelism degree of TP=2 for Qwen3-32B, while setting TP=1 for all other models.

\subsection{Experimental Results}

In this section, we evaluate PARD-2 on Qwen3 and Llama3 with thinking mode disabled. 
We compare PARD-2 with several SD baselines, including EAGLE-3, DFlash, and PARD. For the Qwen3 series, EAGLE-3 uses third-party trained weights~\citep{tencent2025angelslim}, while all other baselines and our method use official weights.  For EAGLE-3, we adopt an inference draft length of $K=8$ to match its optimal open-source configuration, whereas for all remaining methods, we set the draft length to $K=16$.

\textbf{Target-Dependent Mode.} Table~\ref{tab:Target-Dependent Mode} reports the main results under greedy decoding. Across all evaluated target models, PARD-2 consistently outperforms auto-regressive decoding and strong speculative decoding baselines in both speedup and average acceptance length $\tau$. On Qwen3-8B, PARD-2 raises the average speedup to 5.81$\times$, compared with 4.39$\times$ for PARD and 4.61$\times$ for DFlash, while increasing the average acceptance length to 6.98. Similar gains are observed on Qwen3-14B and Llama3.1-8B, where PARD-2 achieves 5.81$\times$ and 5.19$\times$ average speedups, respectively. It is worth noting that PARD-2 maintains strong performance on MT-Bench, which involves more complex multi-turn dialogue generation, suggesting that its benefits generalize beyond structured reasoning and coding benchmarks. These results demonstrate that PARD-2 improves the acceptance of consecutive draft tokens and translates this improvement into consistent lossless inference acceleration in practice.

\textbf{Target-Independent Mode.}
Table~\ref{tab:Target-Independent Mode} evaluates PARD-2 in target-independent mode, where a single drafter accelerates different Qwen target models. Compared with PARD, PARD-2 improves both average speedup and acceptance length. Specifically, PARD-2 increases the average speedup from 4.38$\times$ to 4.82$\times$ on Qwen3-8B, from 4.38$\times$ to 4.79$\times$ on Qwen3-14B, and from 4.37$\times$ to 4.68$\times$ on Qwen3-32B.  On average, the acceptance length $\tau$ improves from 5.41 to 5.97. These results show that stochastic gating reduces over-reliance on target-specific hidden states, while CAT optimization remains effective without target-specific features. Together, they enable a general-purpose drafter to achieve strong lossless acceleration across a family of target models.

\textbf{Large Batch Sizes Study.} We further evaluate PARD-2 under large batch serving settings, where GPU utilization becomes increasingly important. Figure~\ref{fig:bs} reports both per-user throughput (TPS/User) and GPU throughput (TPS/GPU) across different batch sizes. PARD-2 consistently shifts the throughput frontier upward and to the right, indicating that it improves aggregate serving efficiency while maintaining higher per-user generation speed. Notably, even at batch size 64, where the speedup gain is relatively smaller due to higher GPU utilization, PARD-2 still outperforms PARD on both Llama-3-8B and Qwen3-8B. These results show that the gains of PARD-2 are not limited to small-batch; instead, PARD-2 remains effective in high-throughput serving scenarios, where large-batch decoding is commonly used to maximize GPU utilization.

\begin{table*}[!]
    \centering
    
    \begin{minipage}[t]{0.48\linewidth}
        \centering
        \caption{Comparison between fixed-decay and token-adaptive weighting strategies.}
        \label{tab:ab_prob_loss}
        \vspace{0.1cm}
        \resizebox{\textwidth}{!}{
        \begin{tabular}{l cc cc cc cc}
            \toprule
            \multirow{2}{*}{\textbf{Prob Loss}} & \multicolumn{2}{c}{\textbf{HumanEval}} & \multicolumn{2}{c}{\textbf{GSM8K}} & \multicolumn{2}{c}{\textbf{MT-Bench}} & \multicolumn{2}{c}{\textbf{Average}} \\
            \cmidrule(lr){2-3} \cmidrule(lr){4-5} \cmidrule(lr){6-7} \cmidrule(lr){8-9}
            & Speedup & $\tau$ & Speedup & $\tau$ & Speedup & $\tau$ & Speedup & $\tau$ \\
            \midrule
            $\gamma=1.0$       & 5.52 & 7.45 & 4.22 & 5.58 & 2.67 & 3.54 & 4.14 & 5.52 \\
            $\gamma=0.8$       & 5.61 & 7.56 & 4.33 & 5.67 & 2.74 & 3.60 & 4.23 & 5.61 \\
            $\gamma=0.6$       & 4.89 & 6.58 & 3.99 & 5.21 & 2.72 & 3.58 & 3.87 & 5.12 \\
            \textbf{Ours} & \textbf{5.86} & \textbf{7.90} & \textbf{4.43} & \textbf{5.79} & \textbf{2.79} & \textbf{3.67} & \textbf{4.36} & \textbf{5.79} \\
            \bottomrule
        \end{tabular}
        }
    \end{minipage}
    \hfill 
    \begin{minipage}[t]{0.48\linewidth}
        \centering
        \caption{Effect of the stochastic gating ratio for target features.$\rho$ = 0.1 is optimal.}
        \label{tab:ab_mask_ratio}
        \vspace{0.1cm}
        \resizebox{\textwidth}{!}{
        \begin{tabular}{l cc cc cc cc}
            \toprule
            \multirow{2}{*}{\textbf{Gate Ratio}} & \multicolumn{2}{c}{\textbf{HumanEval}} & \multicolumn{2}{c}{\textbf{GSM8K}} & \multicolumn{2}{c}{\textbf{MT-Bench}} & \multicolumn{2}{c}{\textbf{Average}} \\
            \cmidrule(lr){2-3} \cmidrule(lr){4-5} \cmidrule(lr){6-7} \cmidrule(lr){8-9}
            & Speedup & $\tau$ & Speedup & $\tau$ & Speedup & $\tau$ & Speedup & $\tau$ \\
            \midrule
            w/o            & 6.10 & 8.28 & 4.66 & 6.17 & 2.88 & 3.86 & 4.55 & 6.10 \\
            $\rho$ = 0.5          & 5.96 & 8.15 & 4.55 & 6.01 & 2.85 & 3.80 & 4.45 & 5.99 \\
            $\rho$ = 0.2          & 5.99 & 8.18 & 4.66 & 6.12 & 2.85 & 3.80 & 4.50 & 6.03 \\
            \textbf{$\rho$ = 0.1} & \textbf{6.02} & \textbf{8.19} & \textbf{4.68} & \textbf{6.13} & \textbf{2.88} & \textbf{3.85} & \textbf{4.53} & \textbf{6.06} \\
            \bottomrule
        \end{tabular}
        }
    \end{minipage}
    
\end{table*}

\subsection{ABLATION STUDIES}
In this section, we ablate the key design choices of PARD-2, including the effectiveness of CAT optimization, the impact of stochastic gating for target features, and a fine-grained breakdown of the improvements over PARD. All ablation models are trained for 30k steps on MI300X GPUs.

\textbf{Confidence-Adaptive Token Optimization (CAT).} 
CAT prioritizes ``high-value'' tokens that directly extend the accepted prefix during speculative decoding. Unlike traditional uniform supervision, CAT reweights the token-level training loss based on the target model's confidence. As shown in Table~\ref{table:ab_main}, CAT consistently improves the average acceptance length across all benchmarks. With a larger $k_{\mathrm{infer}}$, CAT increases $\tau$ from 4.83 to 5.79 across all benchmarks.

To further validate its superiority, we compare CAT against a fixed position-wise decay strategy~\citep{chen2026dflash, liu2026dart} ($\gamma_t=\gamma^{t-1}$), a common heuristic in parallel drafting. As reported in Table~\ref{tab:ab_prob_loss}, while position-wise decay provides gains (reaching a peak $\tau$ of 5.61 at $\gamma=0.8$), its performance is highly sensitive to the decay rate and fails to generalize across different tasks. In contrast, CAT adaptively focuses on both the token and its prefix. The results demonstrate that incorporating both the token and its prefix into the weighting strategy is essential for achieving optimal speculative decoding.

\textbf{Stochastic Gating for Target Features.}
To balance target-dependent performance and target-independent versatility, we introduce a training-time stochastic gate for target-feature injection. The gate disables target features with probability $\rho$ and injects them otherwise, encouraging the drafter to avoid over-reliance on target hidden states. As shown in Table~\ref{tab:ab_mask_ratio}, the fully injected baseline achieves $\tau=5.62$, while stochastic gating with $\rho=0.1$ maintains a comparable $\tau=5.60$. Notably, $\rho=0.1$ slightly improves MT-Bench performance from 3.79 to 3.84, suggesting that mild stochastic gating acts as an effective regularizer. This helps prevent overfitting to specific target hidden distributions and improves the model's versatility across deployment settings.

\textbf{Analysis of Performance Gains.} 
In Table~\ref{table:ab_main}, we conduct a fine-grained ablation of the new modules in PARD-2, including target-feature injection, CAT, and the draft length. We study conditioned drafting by using target-model hidden representations as additional input features, enabling the draft model to leverage target-side context beyond previous tokens. These features increase the average $\tau$ from 4.70 to 4.96. The gains are larger on reasoning and code-generation tasks, suggesting that target hidden states provide useful semantic signals for resolving complex logic and generating candidates more likely to be accepted.

\begin{table}[!t]
    \centering
    \caption{Ablation study of core components and configurations in PARD-2. Compared to PARD, PARD-2 progressively adds target hidden features, CAT optimization, and multi-layer target features over 
    PARD, and evaluates different draft lengths for training $k_{\text{train}}$ and inference $k_{\text{infer}}$. Each component consistently improves both speedup and average acceptance length ($\tau$) across three benchmarks.}
    \label{tab:ablation_study}
    \vspace{0.3cm}
    \resizebox{\textwidth}{!}{
    \begin{tabular}{l cc cc cc cc cc}
        \toprule
        \multirow{2}{*}{\textbf{Method}} & \multicolumn{2}{c}{\textbf{Draft Length}} & \multicolumn{2}{c}{\textbf{HumanEval}} & \multicolumn{2}{c}{\textbf{GSM8K}} & \multicolumn{2}{c}{\textbf{MT-Bench}} & \multicolumn{2}{c}{\textbf{Average}} \\
        \cmidrule(lr){2-3} \cmidrule(lr){4-5} \cmidrule(lr){6-7} \cmidrule(lr){8-9} \cmidrule(lr){10-11}
        & $k_{train}$ & $k_{infer}$ & Speedup & $\tau$ & Speedup & $\tau$ & Speedup & $\tau$ & Speedup & $\tau$ \\
        \midrule
        PARD                               & 8  & 8  & 4.71 & 5.90 & 3.89 & 4.78 & 2.75 & 3.42 & 3.78 & 4.70 \\
        PARD + Target Feat                 & 8  & 8  & 4.92 & 6.18 & 4.18 & 5.12 & 2.88 & 3.57 & 3.99 & 4.96 \\
        PARD + Target Feat                 & 16 & 8  & 4.76 & 6.00 & 4.11 & 5.02 & 2.80 & 3.46 & 3.89 & 4.83 \\
        PARD + Target Feat + CAT         & 16 & 8  & 4.94 & 6.19 & 4.24 & 5.15 & 2.93 & 3.60 & 4.04 & 4.98 \\
        PARD + Target Feat + CAT          & 16 & 16 & 5.86 & 7.90 & 4.41 & 5.79 & 2.79 & 3.67 & 4.35 & 5.79 \\
        PARD + Multi-layer Target Feat + CAT  & 16 & 16 & 5.89 & 7.93 & 4.41 & 5.80 & 2.86 & 3.79 & 4.37 & 5.84 \\
        \bottomrule
    \end{tabular}
    }
    \label{table:ab_main}
\end{table}

\section{Related Work}

Speculative decoding~\citep{leviathan2023fast, chen2023accelerating} alleviates the memory-bandwidth bottleneck in auto-regressive generation by using a lightweight draft model to propose tokens for parallel verification by a target LLM. To improve draft-target alignment, Medusa~\citep{cai2024medusa}, GLIDE and CAPE~\citep{du2024glide}, and the EAGLE series~\citep{li2024eagle, li2025eagle} incorporate the KV cache or hidden features of the target model, while DistillSpec~\citep{zhou2023distillspec} employs knowledge distillation. To further minimize wall-clock latency, methods such as PEARL~\citep{liupearl} and SSD~\citep{kumarspeculative} decouple drafting and verification for parallel execution, whereas SpecInfer~\citep{miao2023specinfer}, Falcon~\citep{gao2024falcon} and  EAGLE-2~\citep{li2024eagle2fasterinferencelanguage} introduce advanced tree-based verification. Training-free $n$-gram matching methods such as LOOKAHEAD~\citep{Fu2024} and PROMTEC~\citep{lee2025promtec} also accelerate inference.

Despite these advances, many SD methods still rely on auto-regressive drafting, whose sequential dependency limits drafting throughput. Recent parallel drafting methods address this limitation by predicting multiple future tokens in a single forward pass. ParallelSpec~\citep{xiao2024parallelspec} trains a parallel drafter to serve as an efficient speculative model. P-EAGLE~\citep{hui2026p} and PARD~\citep{an2025pard} adapt auto-regressive models to parallel masked prediction, while SpecDiff~\citep{christopher2025speculative}, SpecDiff-2~\citep{sandler2025specdiff}, DART~\citep{liu2026dart} and DFlash~\citep{chen2026dflash} employ diffusion-style drafters for parallel token generation. However, existing parallel methods often rely on uniform token-level supervision. While some approaches~\citep{chen2026dflash, liu2026dart} introduce fixed position-aware decaying weights, they remain suboptimal for aligning with speculative decoding verification. In practice, token acceptance depends on both the prefix context and the token identity, suggesting that supervision weights should be dynamically determined by their joint effect.

\section{Conclusion}
We propose PARD-2, a dual-mode speculative decoding framework that unifies target-dependent and target-independent modes within a single draft model. By analyzing acceptance length in speculative decoding, we identify a gap between training-time objectives and the inference-time goal of maximizing consecutive token acceptance. To bridge this gap, we introduce Confidence-Adaptive Token (CAT) optimization, which uses target-model confidence as a proxy for token-level acceptance and adaptively reweights each token accordingly. Experiments on diverse benchmarks show that PARD-2 improves acceptance length and inference efficiency, demonstrating its effectiveness as a flexible framework for lossless speculative decoding acceleration.
\newpage

\bibliography{neurips_2026}
\bibliographystyle{neurips_2026}

\newpage
\clearpage

{\center \Large \bf \Large{\bf Appendix} \par}

\appendix
\setcounter{section}{0}

\section{Training Hyperparameters}\label{appendix:Hyperparameters}
Table~\ref{tab:training_hyperparameters} summarizes the hyperparameters used for training.
\begin{table}[htbp]
\centering
\caption{Selected Hyperparameters for PARD-2 Training}
\begin{tabular}{l|ccc}
\toprule
\textbf{Hyperparameter} & \textbf{Llama3.1-8B} & \textbf{Qwen3-8B} & \textbf{Qwen3-14B} \\
\midrule
Optimizers & AdamW & AdamW & AdamW \\
\midrule
Learning Rate & 1e-5 & 3e-5 & 3e-5 \\
\midrule
Per Device Train Batch Size & 8 & 4 & 4 \\
\midrule
Gradient Accumulation Steps & 1 & 2 & 2 \\
\midrule
Num Processes & 8 & 8 & 8 \\
\midrule
Num Train Epochs & 4 & 2 & 2 \\
\midrule
Training Draft Length K & 16 & 16 & 16 \\
\midrule
Stochastic Gating Ratio $\rho$ & 0.1 & 0.1 & 0.0 \\
\midrule
CE Loss Coefficient  $\beta$  & 0.1 & 0.1 & 0.1 \\
\midrule
Max Seq Length & 512 & 1024 & 1024 \\

\bottomrule
\end{tabular}
\label{tab:training_hyperparameters}
\end{table}

\end{document}